\documentclass{article}
\usepackage{arxiv}
\usepackage[utf8]{inputenc} 
\usepackage[T1]{fontenc}    
\usepackage{hyperref}       
\usepackage{url}            
\usepackage{booktabs}       
\usepackage{amsfonts}       
\usepackage{amsmath}
\usepackage{amssymb}
\usepackage{nicefrac}       
\usepackage{microtype}      
\usepackage{lipsum}		    
\usepackage{graphicx}
\usepackage{natbib}
\usepackage{doi}
\usepackage{algorithm}
\usepackage{algpseudocode}
\usepackage{upgreek}

\title{On Predictive planning and counterfactual learning in active inference}

\date{\today}

\author{{Aswin Paul\textsuperscript{1, 2, 3}, Takuya Isomura\textsuperscript{4} and Adeel Razi\textsuperscript{1, 5, 6}} \\
    \\
    \textsuperscript{1} Turner Institute for Brain and Mental Health, School of Psychological Sciences, Monash University, Clayton 3800, Australia \\
    \textsuperscript{2} IITB-Monash Research Academy, Mumbai, India \\
    \textsuperscript{3} Department of Electrical Engineering, IIT Bombay, Mumbai, India \\
    \textsuperscript{4} Brain Intelligence Theory Unit, RIKEN Center for Brain Science, Wako, Saitama, Japan \\
    \textsuperscript{5} Wellcome Trust Centre for Human Neuroimaging, University College London, WC1N 3AR London, United Kingdom \\
    \textsuperscript{6} CIFAR Azrieli Global Scholars Program, CIFAR, Toronto, Canada \\
}

\renewcommand{\shorttitle}{On predictive planning and counterfactual learning in AIF}

\begin{document}
\maketitle

\begin{abstract}
Given the rapid advancement of artificial intelligence, understanding the foundations of intelligent behaviour is increasingly important. Active inference, regarded as a general theory of behaviour, offers a principled approach to probing the basis of sophistication in planning and decision-making. In this paper, we examine two decision-making schemes in active inference based on 'planning' and 'learning from experience'. Furthermore, we also introduce a mixed model that navigates the data-complexity trade-off between these strategies, leveraging the strengths of both to facilitate balanced decision-making. We evaluate our proposed model in a challenging grid-world scenario that requires adaptability from the agent. Additionally, our model provides the opportunity to analyze the evolution of various parameters, offering valuable insights and contributing to an explainable framework for intelligent decision-making.
\end{abstract}


\section{Introduction}

Defining and thereby separating the intelligent ``agent'' from its embodied ``environment'', which then provides feedback to the agent, is crucial to model intelligent behaviour. Popular approaches, like reinforcement learning (RL), heavily employ such models containing agent-environment loops, which boils down the problem to agent(s) trying to maximise reward in the given uncertain environment \cite{Sutton2018}.

Active inference has emerged in neuroscience as a biologically plausible framework \cite{Friston2010}, which adopts a different approach to modelling intelligent behaviour compared to other contemporary methods like RL. In the active inference framework, an agent accumulates and maximises the model evidence during its lifetime to perceive, learn, and make decisions \cite{DaCosta2020, Sajid2021, Millidge2020}. However, maximising the model evidence becomes challenging when the agent encounters a highly 'entropic' observation (i.e. an unexpected observation) concerning the agent's generative (world) model \cite{DaCosta2020, Sajid2021, Millidge2020}. This seemingly intractable objective of maximising model evidence (or minimising the entropy of encountered observations) is achievable by minimising an upper bound on the entropy of observations, called variational free energy \cite{DaCosta2020, Sajid2021}.
Given this general foundation, active inference \cite{Friston2017} offers excellent flexibility in defining the generative model structure for a given problem and has attracted much attention in various domains\cite{Kuchling2020, Deane2020}.

In this work, we develop an efficient decision-making scheme based on active inference by combining 'planning' and 'learning from experience'. After a general introduction to generative world models in the next section, we take a closer look at the decision-making aspect of active inference. Then, we summarise two dominant approaches in active inference literature: the first based on planning (Section \ref{dpefe-alg-notes}), and the second based on counterfactual learning (cf. Section \ref{clmethod-gamma}). We compare the computational complexity and data efficiency (cf. Section \ref{performance-comparison}) of these two existing active inference schemes and propose a mixed or hybrid model that balances these two complementary schemes (Section \ref{mixed-model-section}). Our proposed hybrid model not only performs well in an environment that demands adaptability (in Section \ref{performance-mixed-model}) but also provides insights regarding the explainability of decision-making using model parameters (in Section \ref{sec:explainability}).

\section{Methods}

\subsection{Agent-environment loop in active inference}
\label{active-inference-intro}

Generative models are central to establishing the agent-environment loop in an active inference model. The agent is assumed to hold a scaled-down model of the external world that enables the agent to predict the external dynamics and future observations. The agent can then use its available actions to pursue future outcomes, ensuring survival. We stick to a partially observed Markov decision process (POMDP) based generative model \cite{Kaelbling1998} in this paper. POMDPs are a general case of Markov decision processes (MDPs), which are controllable Markov chains apt for modelling stochastic systems in a discrete state space. In the following section, we provide the specific details of a POMDP-based generative model.

\subsection{POMDP-based generative models}
\label{pomdp-notes}

In active inference, agents learn the generative model about external states and optimise their decisions by minimising variational free energy. The POMDP is a universal framework to model discrete state-space environments, where the likelihood and state transition are expressed as tractable categorical distributions. Thus, we adopted the POMDP as our agent's generative model. The POMDP-based generative model is formally defined as a tuple of finite sets $(S, O, T, U, \mathbb{B},\mathbb{A}, \mathbb{D}, \mathbb{E})$ such that:

\begin{itemize}
    \item [$\circ$] $s_t \in S:$ states and $s_{1}$ is a given initial state.
    \item [$\circ$]$o_t \in O:$ where $o_t=s_t$ in the fully observable setting, and $o_t=f(s_t)$ in a partially observable setting.
    \item [$\circ$] $T\in \mathbf{N}^{+}$, is a finite time horizon available per episode. 
    \item [$\circ$] $u_t \in U:$ actions, for e.g., $U$=\{\text{Left, Right, Up, Down}\}. 
    \item [$\circ$]$\mathbb{B}:$ encodes one-step transition dynamics, such that $P(s_{t} \vert s_{t-1}, u_{t-1},\mathbb{B})$ is the probability that action $u_{t-1}$ taken at state $s_{t-1}$ at time $t-1$ results in $s_{t}$ at time $t$.
    \item [$\circ$] $\mathbb{A}:$ encodes the likelihood distribution, $ P(o_t \vert s_t, \mathbb{A})$ for  the partially observable setting.
    \item [$\circ$]$\mathbb{D}:$ prior about the state ($s$) at the starting time point used for the Bayesian inference of state ($s$) at time $t=1$.
    \item [$\circ$]$\mathbb{E}:$ prior about action-selection used to take action in the simulations at time $t=1$.
\end{itemize}

In the POMDP, hidden states ($s$) generate observation ($o$) through the likelihood mapping ($\mathbb{A}$) in the form of a categorical distribution, $P(o_t \vert s_t, \mathbb{A}) = \mathrm{Cat}(\mathbb{A})$. The states $s$ are determined by the transition matrix ($\mathbb{B}$) given the agent's action ($u$), $P(s_{t} \vert s_{t-1}, u_{t-1},\mathbb{B}) = \mathrm{Cat}(\mathbb{B}(s_{t-1} \otimes u_{t-1}))$. Thus, the generative model in question is given as:

\begin{equation}
P(o_{1:t},s_{1:t},u_{1:t}) = P(\mathbb{A}) P(\mathbb{B}) P(\mathbb{D}) P(\mathbb{E}) \prod_{\tau=1}^t P(o_{\tau} \vert s_{\tau}, \mathbb{A}) \prod_{\tau=2}^t P(s_{\tau} \vert s_{\tau-1}, u_{\tau-1},\mathbb{B}).
\end{equation}

Under the mean-field approximation, an approximate posterior distribution (concerning hidden-states $s$) is given as:
\begin{equation}
\label{beliefspomdp}
\underbrace{Q(s_{t+1})}_\text{Posterior} = \sigma \left(\underbrace{\text{log}P(s_{t+1})}_\text{Prior} + \underbrace{\text{log}(o_{t+1} \cdot \mathbb{A} s_{t+1})}_\text{Likelihood} \right),
\end{equation}

where the posterior beliefs about states and parameters are expressed as categorical distribution, $Q(s_t) = \mathrm{Cat}(\mathbf{s}_t)$ and Dirichlet distribution, $Q(\mathbb{A}) = \mathrm{Dir}(\mathbf{a})$, respectively. Hence, under this POMDP setup, variational free energy is given as:

\begin{align}
\label{eq:vfe}
F = \sum_{s_{1:t}} Q(s_{1:t})[~\text{log} Q(s_{1:t})-\text{log} P(o_{1:t}\vert s_{1:t}) - \text{log}P(s_{1:t})~] \nonumber\\
+ \sum_{u_{1:t}} Q(u_{1:t})[~\text{log} Q(u_{1:t})-\text{log} P(u_{1:t}\vert s_{1:t})~] + D_{\mathrm{KL}}~[Q(\theta)||P(\theta)].
\end{align}

Variations of $F$ give appropriate posterior expectations about states and parameters. Some optional parameters, depending on the specific decision-making scheme used, are:

\begin{itemize}
    \item [$\circ$]$\mathbb{C}:$ prior preferences over outcomes, $P(o|\mathbb{C})$. Here, $\mathbb{C}$ is the preference for the predefined goal state. This parameter is generally used in the planning-based active inference models \cite{Sajid2021, Paul2021}.
    \item [$\circ$]$\Gamma(t)$: A time-specific risk parameter that the agent maintains to update the state-action mapping $\mathbb{CL}$ in the CL scheme as in \cite{Isomura2022a}.
    \item [$\circ$]$\beta(s,t)$: A state-dependent bias parameter used in the mixed model proposed in this paper.
\end{itemize}

These are used to parameterise the distribution of actions $u$, and actions are optimised through variational free energy minimisation. Further details are explained in the subsequent sections.

\subsection{Decision-making schemes in active inference}
\label{decision-making-aif}

Decision-making under active inference is formulated as minimising the (expected) variational free energy of future time steps \cite{Kaplan2018, Friston2009, Friston2012}. This enables an agent to deploy a planning-based decision-making scheme, where an agent predicts possible outcomes and makes decisions to attain states and observations that minimise expected free energy (EFE). Classically, active inference optimises policies -- i.e., sequences of actions in time -- instead of a state-action mapping in methods like Q-Learning \cite{Sutton2018} in RL to choose the policy that minimises EFE \cite{Sajid2021}. However, such formulations limit agents to solve environments only with low-dimensional state-space \cite{Sajid2021, Paul2021}.

Several improvements to the framework followed, including the recent sophisticated inference scheme \cite{Friston2021} that uses a recursive form of free energy to ease the computational complexity of policy search. The sophisticated inference method uses a forward tree search in time to evaluate EFE; however, it restricts the planning depth of agents \cite{Friston2021} due to computational complexity. More innovative algorithms like dynamic programming can be used to linearise the planning \cite{Paul2023, DaCosta2020}. The proposed linearised planning method was called Dynamic programming in expected free energy (DPEFE) in \cite{Paul2023}. This DPEFE algorithm performs at par with benchmark reinforcement learning methods like Dyna-Q in environments similar to grid world tasks \cite{Paul2021} (See Section \ref{dpefe-alg-notes} for technical details of this method). A generalisation of the DPEFE algorithm was recently proposed as `inductive-inference' to model 'intentional behaviour' in agents \cite{Friston2023ActiveIA}. 

Another recent work deviates from this classical approach of predictive planning and employs ``learning from experience'' to determine optimal decisions \cite{Isomura2022a}. This scheme is mathematically equivalent to a particular class of neural networks accompanied by some neuromodulations of synaptic plasticity \cite{Isomura2020, Isomura2022a}. It uses counterfactual learning (the CL method in this paper) to accumulate a measure of `risk' over time-based on environmental feedback. Subsequent work that validates this scheme experimentally using in-vitro neural networks has also appeared recently \cite{Isomura2023}.

The following summarises the critical algorithmic details of both schemes: DPEFE in Sec.\ref{dpefe-alg-notes} and CL scheme in Sec.\ref{clmethod-gamma}. Both schemes are proposed based on conventional POMDPs. 

\subsubsection{DPEFE scheme and action precision}
\label{dpefe-alg-notes}

The DPEFE scheme in this paper is based on the work in \cite{Paul2021}. This scheme was generalised to a POMDP setting in the paper \cite{Paul2023}. The model parameters used are as given in Sec.\ref{pomdp-notes}. The action-perception loop in the DPEFE scheme comprises perception (i.e., identifying states that cause observations), planning, action selection, and learning model parameters. In this paper, all environments are fully observable since our focus is on decision-making rather than perception, hence $O = S$.

The action selection in the DPEFE scheme is implemented as follows: 
After evaluating the expected free energy (EFE, $\mathbb{G}$) of future observations using dynamic programming (cf. \cite{Paul2023}), the agent evaluates the probability distribution for selecting an action $u$ as: 

\begin{equation}
    P_\mathrm{DPEFE}(u \vert s) = \sigma \left( - \alpha ~ \mathbb{G} \left( u \vert s \right) \right).
    \label{dpefe-action-selection-eqn}
\end{equation}

Here, $\sigma$ is the classical softmax function, rendering actions with smaller EFE being selected with larger probabilities. The action precision parameter ($\alpha$) may be tuned to increase/decrease the agent's action selection confidence. For a detailed description of the evaluation of the EFE ($\mathbb{G}$) and the DPEFE algorithm, refer to \cite{Paul2023} (Section 5).

\subsubsection{CL method and risk parameter}
\label{clmethod-gamma}

Instead of attempting to minimise the EFE directly, in the counterfactual learning (CL) method, the agent learns a state-action mapping $\mathbb{CL}$. This state-action mapping is learned through an update equation mediated by a 'risk' term $\Gamma_{t}$ as defined in \cite{Isomura2022a}:

\begin{equation}
    \mathbb{CL} \leftarrow \mathbb{CL} + t ~ \langle~ (1 - 2 ~\Gamma_{t}) \langle u_{t} \otimes s_{t-1} \rangle ~\rangle.
    \label{cp-clmodeleqn}
\end{equation}

Here, $\langle \cdot \rangle$ refers to the average over time, and $\otimes$ is the Kronecker-product operator.
Given the state-action mapping $\mathbb{CL}$, the agent samples actions from the distribution, 
\begin{equation}
    P(u \vert s)_{CL} = \sigma \left( \ln~ \mathbb{CL} \cdot s_{t-1} \right).
    \label{cl-action-selection-eqn}
\end{equation}

In the simulations, $\Gamma_{t}$ with the following functional form is used: When the agent is at the start position --- or when the agent's action causes a ``high risk'' --- the value of 0.9 is substituted, i.e., $\Gamma_{t} \leftarrow 0.9.$
Otherwise, $\Gamma_{t}$ decreases continuously following the equation
\begin{equation}
    \Gamma_{t} \leftarrow \Gamma_{t} - \frac{1}{T_{goal} - t}.
    \label{gamma-update-eqn}
\end{equation}

Here, $T_{goal}$ is when the agent receives a positive environmental reward. So, the sooner the agent comes to the desirable state, the quicker the $\Gamma_{t}$ (i.e., risk) converges to zero \footnote{For the exact form of the generative model and free energy, refer to \cite{Isomura2020}.}.

All the update rules defined in the paper can be derived from the postulate that the agent tries to minimise the (variational) free energy (Eq. \ref{eq:vfe}) w.r.t the generative model \cite{Paul2023, Isomura2022a}. In the rest of the paper, we investigate the performance of the two schemes --- i.e. the DPEFE and the CL method --- and consider a scheme combining them. The following section explores how these two schemes perform in a given environment. 

\section{Results}

\begin{figure}
  \centering  
  \includegraphics[width=\textwidth, page =
  3]{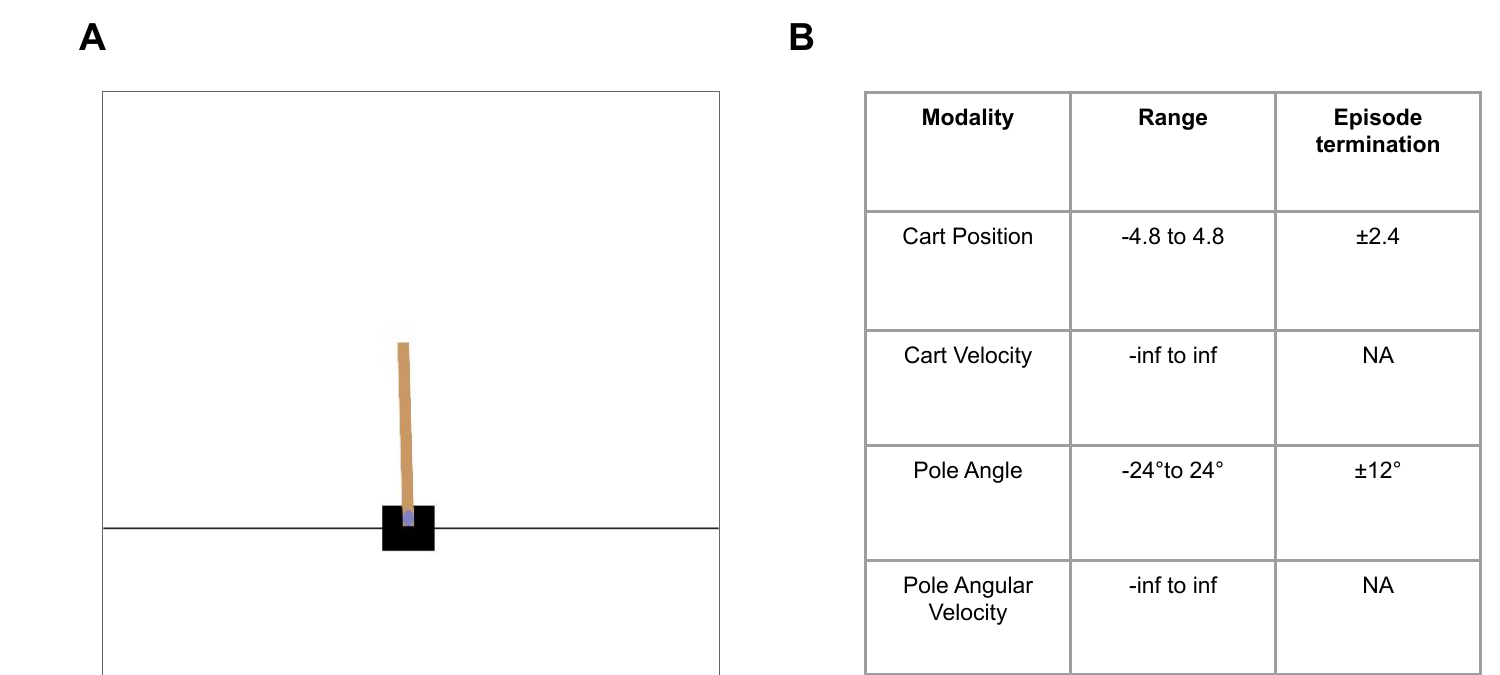}
  \caption{\textbf{A:} A snapshot from the Cart Pole - v1 environment (from OpenAI Gym), \textbf{B:} Environment summary: The objective is to balance the pole (brown) upright as long as possible without meeting the episode termination criteria, i.e. without the pole and cart crossing pole-angle and cart-position thresholds respectively.}
  \label{cartpole}
\end{figure}

We now test the performance of two decision-making schemes (DPEFE and CL) in benchmark environments such as the Cart Pole - v1 (Fig. \ref{cartpole}) from OpenAIGym.

\subsection{Cart Pole - v1 (OpenAI Gym task)}

\begin{figure}
  \centering  
  \includegraphics[width=\textwidth, page =4]{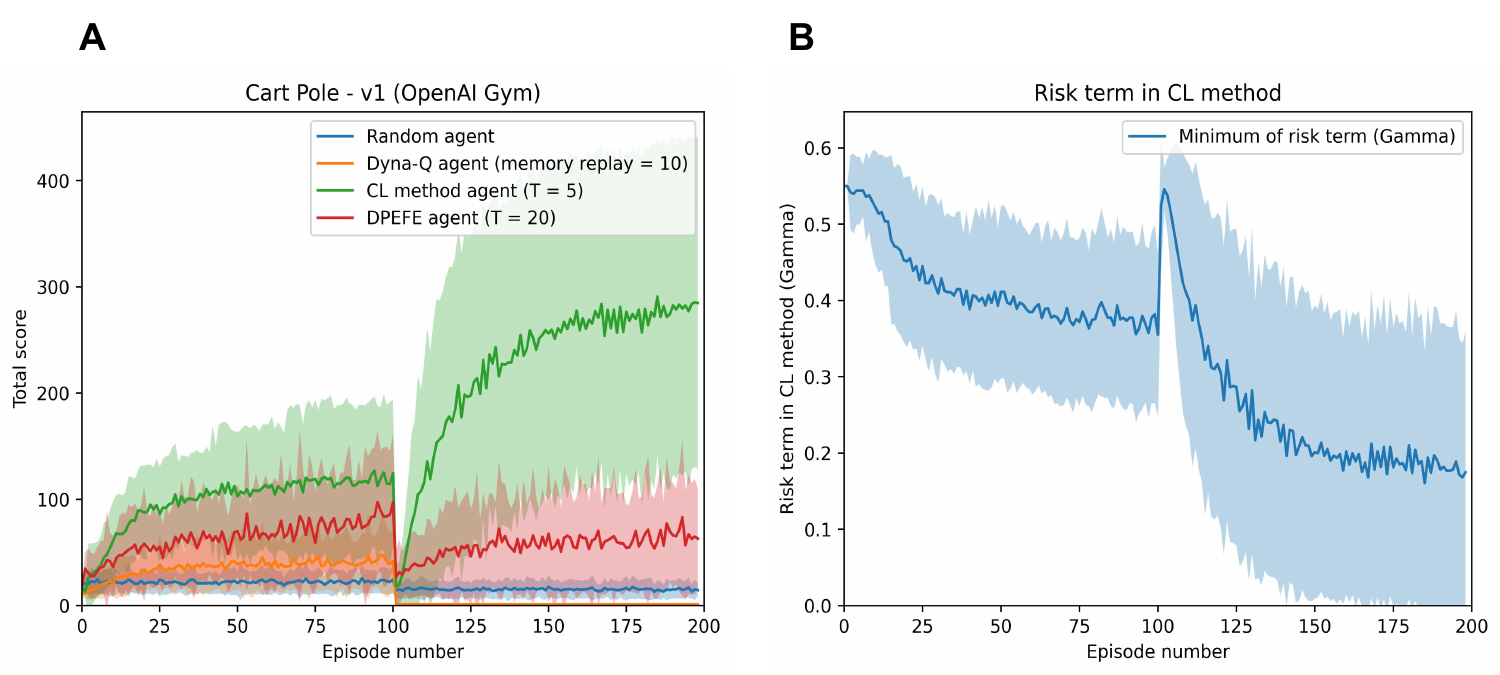}
  \caption{\textbf{A:} Performance of active inference agents with different decision-making schemes in the mutating Cart Pole - v1 ( with a mutation at episode 100). After Episode 100, the environment mutates to a harder version, which the agents must adapt to. \textbf{B:} Evolution of the risk parameter ($\Gamma_t$) of the CL method agent when embodied in the Mutating Cart Pole problem. We can observe the spike at episode 100, consistent with mutation, and the reduced risk resulting in improved performance in the second half of the trial.}
  \label{perf_cartpole}
\end{figure}

In a Cart Pole - v1 environment, an agent is rewarded for balancing the pole upright (within an acceptable range) by moving the cart sideways (Fig.\ref{cartpole} (A)). An episode terminates when the pole or cart crosses the acceptable range ($\pm$12 degrees for the pole and $\pm$2.4 units frame size for the cart, Fig.\ref{cartpole} (B)). This problem is inherently spontaneous, without the need for planning from the controller, where the agent must react to the current situation of the cart and the pole.

We then test the active inference in a mutating setup, where the environment mutates to a more challenging version with half the acceptable range for both the pole and cart position ($\pm$6 degrees for the pole and $\pm$1.2 units frame size for the cart). The performance of the active inference agents with different planning is summarised in Fig.\ref{perf_cartpole} (A).

As expected, the CL method agent outperforms other active inference schemes (As the problem demands spontaneous control, favouring a state-action mapping over planning). The agents quickly learn the necessary state-action mapping and balance the pole more effectively than other planning-based schemes. We observe this also after the mutation in the environment at episode number 100. The improved performance of the CL method agent after mutation warrants additional investigation; however, it can be attributed to the increased feedback frequency due to the increased failure rate after mutation.

In Fig.\ref{perf_cartpole} (B), we see the evolution of the risk term ($\Gamma$). The risk $\Gamma$ settles to a value less than $0.5$ as the agent learns more about the environment. It is interesting to note the increase in $\Gamma$ when faced with a mutation in the environment in Fig.\ref{perf_cartpole} (B) as expected. The risk-reducing behaviour correlates with the increase in performance after episode number 100, highlighting the explainability of the active inference framework.
Next, we test the agents in a fundamentally different environment -- a maze task -- which warrants the need for planning for the future.

\subsection{Complex maze task and data-complexity trade-off}
\label{performance-comparison}

\begin{figure}[tb]
  \centering
  \includegraphics[width=\textwidth, page = 1]{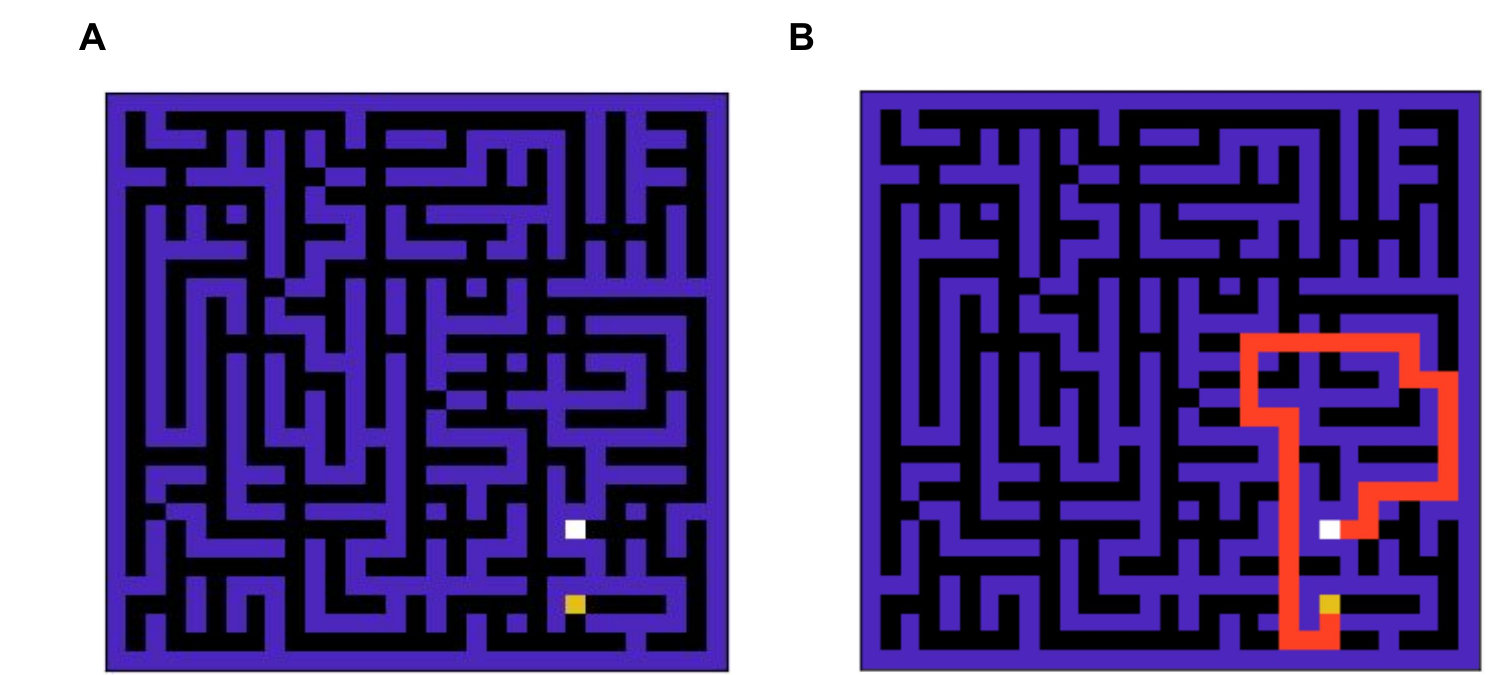}
  \caption{\textbf{A:} A snapshot of the 900-state grid world (maze) environment. \textbf{B:} The optimal solution for the maze is shown in \textbf{A}. This is a complex maze, as when actions are taken randomly, it takes around $9000$ steps to navigate the grid against the optimal route with $47$ steps.}
  \label{gridandsolution}
\end{figure}

To compare the performance of the two agents in a strategic task, we simulate the performance in a standard grid world task as shown in Fig.\ref{gridandsolution} (A). The optimal solution to this grid problem is demonstrated in Fig.\ref{gridandsolution} (B). This is a complex grid world, which is non-trivial compared to grid world tasks used in the past literature to solve \cite{Sajid2021}, as it will take around nine thousand steps for an agent to reach the goal state if actions are taken randomly against the optimal route with length $47$.

\begin{figure}
  \centering
  \includegraphics[width=\textwidth, page = 2]{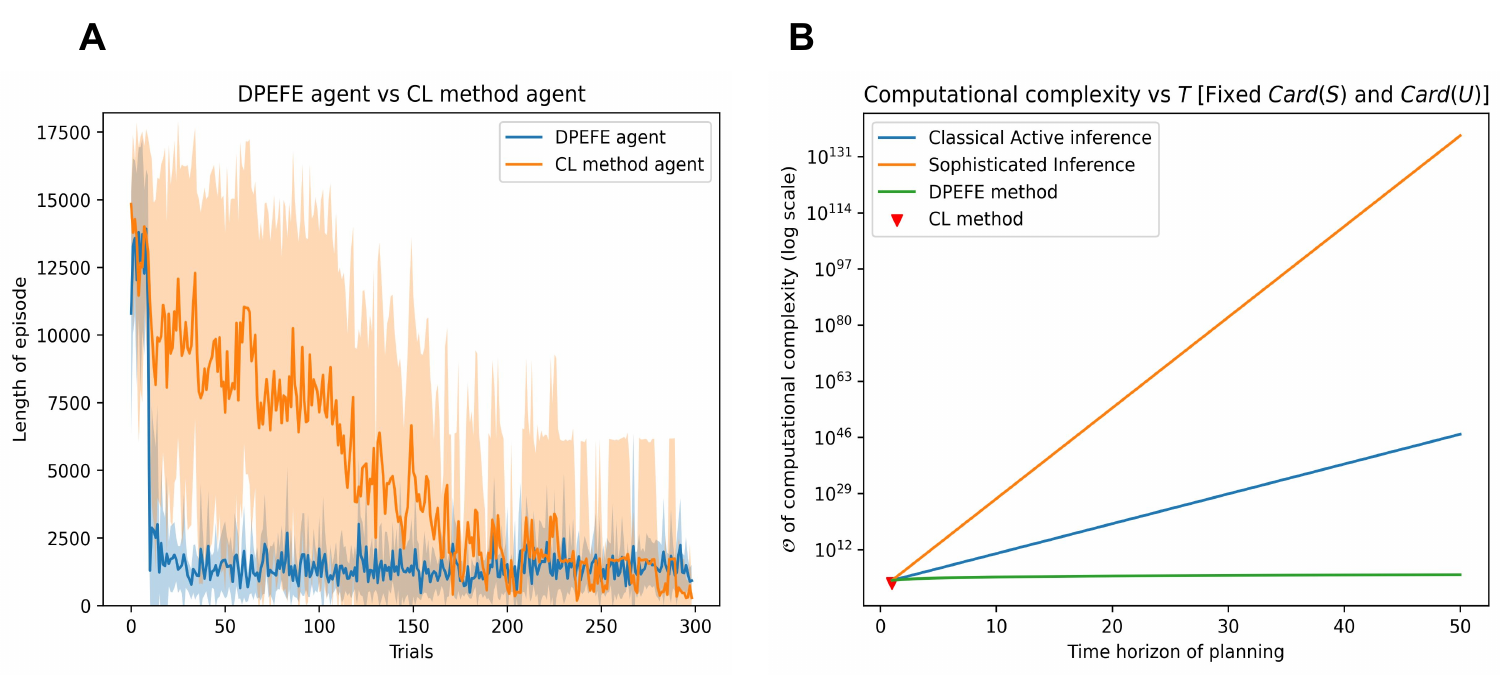}
  \caption{\textbf{A:} Performance comparison of DPEFE and CL agents in the 900-state grid scheme with 300 episodes. The DPEFE agent learns to navigate the grid faster (With a lower episode length) than the CL method agent. \textbf{B:} Comparison of computational complexity between state-of-the-art active inference algorithms \cite{Sajid2021, Friston2021}, the DPEFE method \cite{Paul2021} and CL method \cite{Isomura2022a}. Please note that the y-axis is in the log scale. The computational complexity was calculated for the algorithms to implement planning in a standard grid like in Fig.\ref{gridandsolution}.}
  \label{dc-tradeoff}
\end{figure}

The performance is evaluated regarding how soon the agent can finish an episode (i.e., the length of an episode (lower the better) for reaching the goal state). The simulation results showing the performance of DPEFE and CL agents are plotted in Fig.\ref{dc-tradeoff} (A). These results show that the predictive planning-based DPEFE agent can learn quickly (i.e., within ten episodes) to navigate this grid. In the simulations, the action precision used by the DPEFE agent is $\alpha = 1$ substituted in \eqref{dpefe-action-selection-eqn}. The agent tends to navigate in even lower time steps for a higher action precision ($\sigma$), always sticking to optimal actions. Additionally, we observe that the CL method agent takes longer to learn the optimal path. This result (Fig.\ref{dc-tradeoff} (A)) shows that the CL agent needs more experience in the environment (i.e. more data) to solve it.

In Fig.\ref{dc-tradeoff} (B), we compare major active inference algorithms' computational complexity associated with planning for decision-making. The DPEFE algorithm is computationally efficient compared to other popular active inference schemes \cite{Sajid2021, Friston2021}. Please note that this figure also emphasises how the CL method has no computational complexity associated with planning. So, it is clear that the CL method agent is computationally cheaper than the DPEFE agent as there is no planning component. The computational complexity of the DPEFE agent is associated with the planning depth (time horizon of planning, $T$), as seen in Fig.\ref{dc-tradeoff} (B). This demonstrates a data-complexity trade-off between both these schemes. 

This realisation motivates us towards a mixed model, where we propose to develop an agent that can balance the two schemes according to the resources available to the agent. This makes much sense from the neuro-biological perspective, as biological agents continually try to balance resources to learn and plan for the future versus the experience they already have. This idea also relates to the classic exploration-exploitation dilemma in reinforcement learning \cite{Triche2022}.

\subsection{Integrating the two decision-making approaches}
\label{mixed-model-section}

\begin{figure}
  \centering
{\includegraphics[width = \textwidth, page=1]{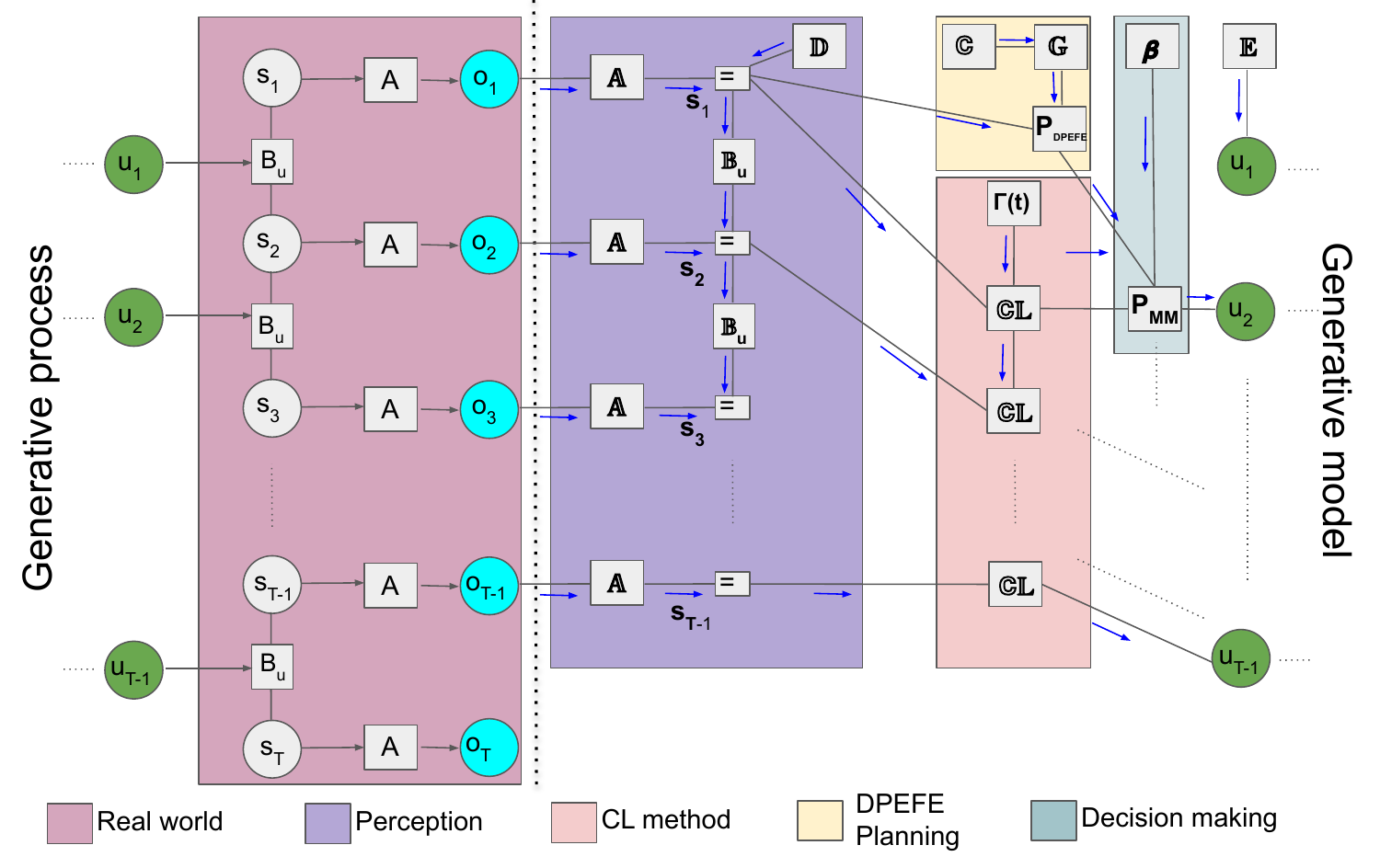}}
  \caption{Flow diagram of the agent-environment loop in the proposed mixed model combining planning and counterfactual learning. There is a key distinction between the generative process and the generative model in the active inference framework. In a POMDP, we assume that the observations are generated by the generative process (`Real world') by `hidden-states' ($s_t$) through a state-observation mapping ($A$), both being inaccessible to the agent. In the generative model, the agent uses $o_t$ to maintain an optimal belief about the hidden state $s_t$ (`Perception'). Subsequently, the agent uses the planning method (DPEFE) and Counterfactual (CL) method to combine the action distributions using the model-bias parameter $\beta$ for decision-making. The decision at time $t$ influences the hidden state of the `Real world' at the next step, completing the loop. The generative process can be thought of as the environment the agent tries to survive in, whereas the generative model is completely part of the agent and can be interpreted as the `imaginary' world the agent assumes it survives.}
  \label{mmmodel}
\end{figure}

To enable the agent to balance its ability to predict future outcomes and use prior experience, we introduce a state-dependent bias parameter that evolves with experience ($\beta (s,t) ~\in ~\left[0,1\right]$) to the model. This addition is motivated by the hypothesis that an agent maintains a sense of bias, quantifying its confidence in the experience of deciding (in the past) in that particular state.

When exposed to a new environment, an agent starts with an equal bias for DEEFE (predictive planning) and CL schemes, represented by a prior bias parameter $\beta_{\text{prior}} = 0.5$.

Over the episodes, the agent will have the probability distributions for decision-making from both models. These distributions enable decision-making given the present state ($s$). In a fully observable environment (MDP), $s$ is known to the agent (i.e. $O$ = $S$, or $\mathbb{A} = \mathbb{I}$, the identity mapping). In the partially observable case (POMDP), the agent infers the (hidden) state ($s$) from observation ($o$) by minimising variational free energy \cite{DaCosta2020, Sajid2021}.

Given the state estimation, $P(u \vert s)_{\text{DPEFE}}$ and $P(u \vert s)_{\text{CL}}$ are the distributions used for sampling decision-making corresponding to the DPEFE scheme and CL method respectively (See Section \ref{dpefe-alg-notes} and Section \ref{clmethod-gamma} for details).

Given these distributions, the agent can now evaluate how 'useful' they are using their Shannon entropy ($\mathbb{H}$(X)). This measure is beneficial as it represents how 'sure' that particular distribution is regarding a decision in that state(s). Namely, if the agent has confidence in a specific action, the action distribution tends to be a one-hot vector favouring the confident action; hence, the entropy of the distribution tends to zero, in contrast to the uniform distribution (not favouring any action) with maximum entropy. Thus, comparing this quantity enables the selection of the most confident strategy from the pool of different schemes.

Based on this observation, over time, the agent can use this entropy measure to update the value of $\beta (s,t)$ as follows:
\begin{equation}
    \beta(s_t) \leftarrow \beta(s_t) + \alpha \left(\mathbb{H}(P_{\text{CL}}(u \vert s_t)) -  \mathbb{H}(P_{\text{DPEFE}}(u \vert s_t)) \right).
    \label{beta-update-eqn}
\end{equation}

Here, $\alpha$ is a normalisation parameter stabilising the updated value, and we make sure that $\beta ~\in ~\left[0,1\right]$ by re-calibrating $\beta < 0$ as $\beta = 0$ and $\beta > 1$ as $\beta = 1$. From a Bayesian inference perspective, one may view the updated belief $\beta$ in Eq.\eqref{beta-update-eqn} as a posterior belief representing how likely the DPEFE model is selected, similar to the Bayesian model selection schemes.

Using this measure of bias $\beta(s_t)$, the agent can now evaluate a new distribution for decision-making, $P_{MM}$, where MM stands for the mixed model as:

\begin{equation}
    P(u \vert s_t)_{\text{MM}} = P(u \vert s_t)_{\text{CL}}^{1-\beta(s_t)} \cdot P(u \vert s_t)_{\text{DPEFE}}^{\beta(s_t)}.
    \label{pmm-eqn}
\end{equation}

The flow diagram describing the proposed mixed model's POMDP-based ``agent-environment'' loop is given in Fig.\ref{mmmodel}. \footnote{For a detailed description of various parameters in the hybrid model, refer to Section \ref{pomdp-notes}, Section \ref{dpefe-alg-notes}, and Section \ref{clmethod-gamma}.}.

\subsection{Deriving update equations for the mixed model from variational free energy}

Eqs.\ref{beta-update-eqn} and \ref{pmm-eqn} can be derived from variational free energy minimisation under a POMDP generative model. The variational free energy for the mixed model is defined as:

\begin{align}
F = \sum_{\tau=1}^t \mathbf{s}_{\tau} \cdot \left\{ \ln \mathbf{s}_{\tau} - \ln \mathbb{A} \cdot o_{\tau} - \ln \mathbb{B}~ \mathbf{s}_{\tau-1} \right\} \nonumber\\
+ \sum_{\tau=1}^t \mathbf{u}_{\tau} \cdot \left\{ \ln \mathbf{u}_{\tau}  + \boldsymbol{\upbeta} \mathbf{\alpha} \cdot \mathbb{G}_{\tau} - (1-\boldsymbol{\upbeta}) (1-2\Gamma_t)\ln \mathbb{CL} ~\mathbf{s}_{\tau-1} \right\} \nonumber\\
+ D_{\mathrm{KL}}~[Q(\beta)||P(\beta)] + D_{\mathrm{KL}}~[Q(\theta)||P(\theta)]
\end{align}

When $\Gamma_t=0$ and $\beta_{\text{prior}} = 0.5$, the derivative of $F$ with respect to $\boldsymbol{\upbeta} = \mathrm{E}[\beta]$ gives the posterior expectation as follows:

\begin{equation}
\boldsymbol{\upbeta} = \mathrm{sig}\left(-\sum_{\tau=1}^t \mathbf{u}_{\tau} \cdot \mathbf{\alpha} \cdot \mathbb{G}_{\tau} - \sum_{\tau=1}^t \mathbf{u}_{\tau} \cdot \ln \mathbb{CL} \mathbf{s}_{\tau-1}\right).
\end{equation}

Interestingly, this posterior expectation can be rewritten using the entropies of DPEFE and CL. The above $F$ becomes variational free energy (Eq. \ref{eq:vfe}) for DPEFE or CL when $\beta=1$ or 0, respectively. 

Thus, minimising $F$ with respect to $\mathbf{u}_{\tau}$ yields:
\begin{equation}
    \mathbf{u}_{\tau}=\sigma(-\mathbf{\alpha} \cdot \mathbb{G}_{\tau}),
\end{equation}
for DPEFE and,
\begin{equation}
    \mathbf{u}_{\tau}=\sigma(\ln \mathbb{CL} \mathbf{s}_{\tau-1}),
\end{equation}
for CL (note that $\Gamma_t=0$ is usually supposed in CL when generating actions).

Thus, from the definition of the Shannon entropy, we obtain
\begin{equation}
\mathbb{H}_{\mathrm{DPEFE}} = -\sum_{\tau=1}^t \mathbf{u}_{\tau} \cdot \ln \mathbf{u}_{\tau} = \sum_{\tau=1}^t \mathbf{u}_{\tau} \cdot \mathbf{\alpha} \cdot \mathbb{G}_{\tau},
\end{equation}
and, 
\begin{equation}
    \mathbb{H}_{\mathrm{CL}} = -\sum_{\tau=1}^t \mathbf{u}_{\tau} \cdot \ln \mathbf{u}_{\tau} = -\sum_{\tau=1}^t \mathbf{u}_{\tau} \cdot \ln \mathbb{CL} \mathbf{s}_{\tau-1}.
\end{equation}

Hence, $\boldsymbol{\upbeta}$ can be rewritten as:
\begin{equation}
\boldsymbol{\upbeta} = \mathrm{sig}(-\mathbb{H}_{\mathrm{DPEPE}} + \mathbb{H}_{\mathrm{CL}}).
\label{beta-update-eqn2}
\end{equation}

When $|\mathbb{H}_{\mathrm{CL}} - \mathbb{H}_{\mathrm{DPEPE}}| << 1$, Eq.\ref{beta-update-eqn} approximates Eq.\ref{beta-update-eqn2}. Minimisation of $F$ further yields Eq.\ref{pmm-eqn} as it is an expression using the probability distribution and equivalent to the posterior expectation:
\begin{equation}
\mathbf{u}_{\tau}=\sigma(-\boldsymbol{\upbeta} \mathbf{\alpha} \cdot \mathbb{G}_{\tau} + (1-\boldsymbol{\upbeta}) \ln \mathbb{CL} \mathbf{s}_{\tau-1}). 
\end{equation}

Therefore, the update rules for the mixed model (Eqs.\ref{beta-update-eqn} and \ref{pmm-eqn}) can be formally derived from variational free energy minimisation.

\subsection{Performance of mixed-model in a mutating maze environment}
\label{performance-mixed-model}

\begin{figure}[tpb]
  \centering
  \includegraphics[width=\textwidth, page = 5]{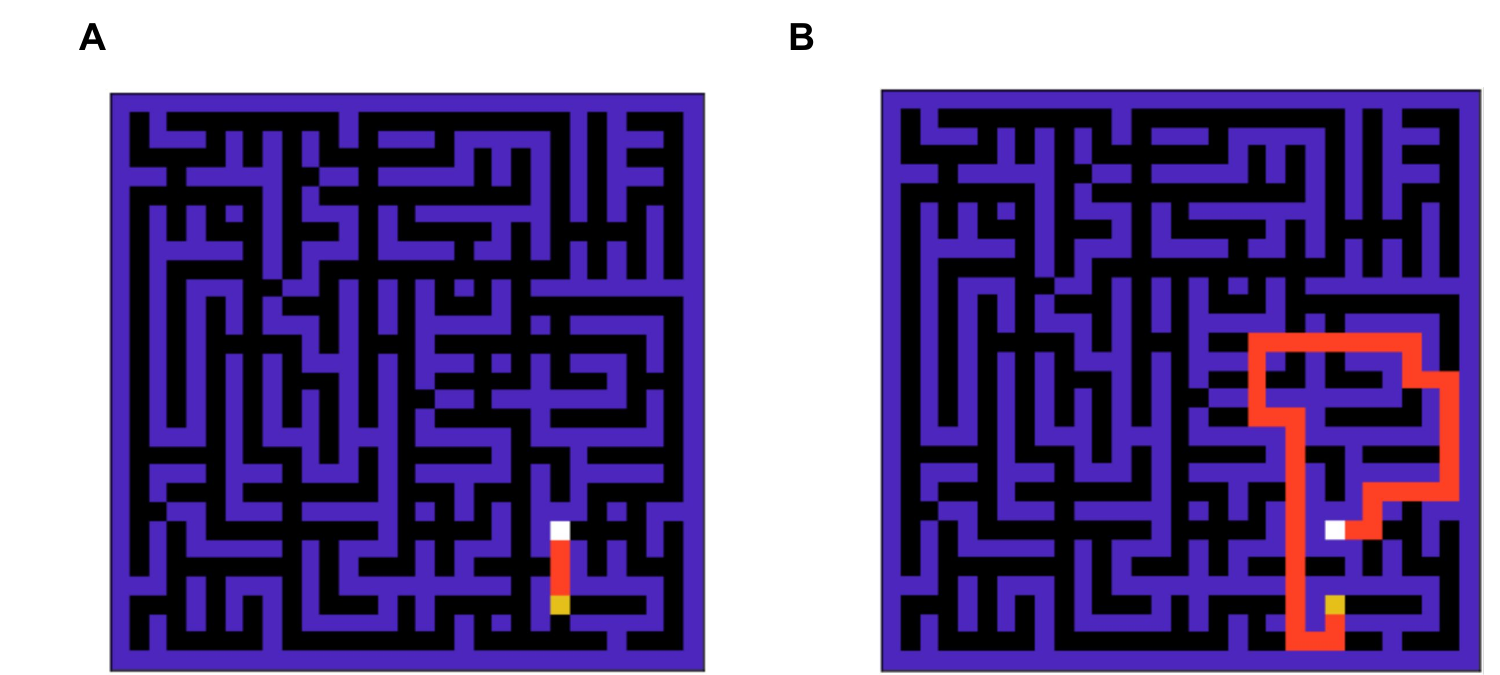}
  \caption{The mutating grid scheme used for studying agent's adaptability. The agent learns to navigate the easy grid (\textbf{A}) in the first half (300 episodes) and faces environment mutation and should learn to solve the hard grid (\textbf{B}).}
  \label{mutatinggridsetup}
\end{figure}

\begin{figure}[tpb]
  \centering 
  \includegraphics[width=\textwidth, page = 6]{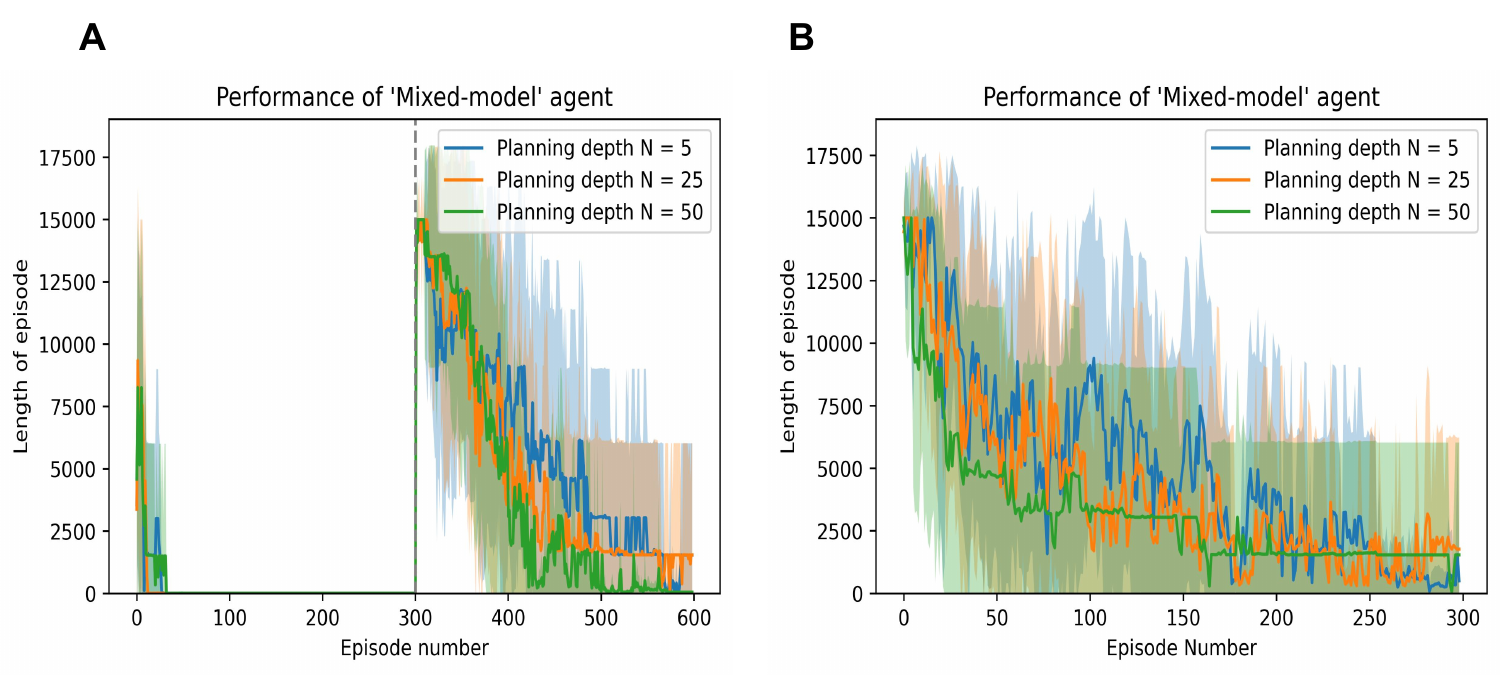} 
  \caption{\textbf{A:} Performance of mixed model agents with different planning depths in the mutating grid scheme, \textbf{B:} Performance of mixed model agents with different planning depths in the hard maze simulated separately.}
  \label{mixed-model-performance}
\end{figure}

We now examine the proposed mixed scheme with agents of different planning power (i.e. different planning depths, $N$ \footnote{We refer to the planning horizon of the mixed-model as $N$, and the DPEFE method as $T$ to avoid confusion.}) in a similar environment. The computational complexity of the DPEFE scheme is linearly dependent on the planning time horizon (planning depth), i.e. $T$, and holds for the mixed-model agent as well (see Fig.\ref{dc-tradeoff}). Thus, an agent with planning depth $N=50$ takes up twice the computational resources while planning compared to an agent with $N=25$.

We use a mutating grid environment to test the performance of the mixed model-based agent. This mutating grid scheme is illustrated in Fig.\ref{mutatinggridsetup}. The agent starts in a more accessible grid version with an optimal path of four steps (Fig.\ref{mutatinggridsetup}, (A)). After 300 episodes, the environment mutates to the complex version of the grid shown in the previous section (See Fig.\ref{mutatinggridsetup} (B)). This setup also enables us to study how adaptable the agent is to new environmental changes.

The performance is summarised in Fig.\ref{mixed-model-performance}. We observed that all three mixed model agents (with varying levels of planning ability) learned to navigate the more accessible grid within the first ten episodes (Fig.\ref{mixed-model-performance}: A). However, when the environment mutated to the rigid grid in episode number 300, the agents learned similar to the performance we observe when navigating that grid alone, Fig.\ref{mixed-model-performance}: B, (i.e., complex grid with 900 states).

We also observed that the agent with higher planning ability learned to navigate the grid faster and more confidently than the other two. This result demonstrates that the proposed mixed model enables agents to balance the two decision-making approaches in the active inference framework.

It is considered that the brain of biological organisms also employs mechanisms to switch multiple strategies. Our model is potentially helpful for unveiling efficient decision-making mechanisms in the brain and their neuronal substrates and developing computationally efficient bio-mimetic agents.

\section{Discussion}

\subsection{Explainability of the active inference models}
\label{sec:explainability}

\begin{figure}
  \centering
  {\includegraphics[width=\textwidth, page = 7]{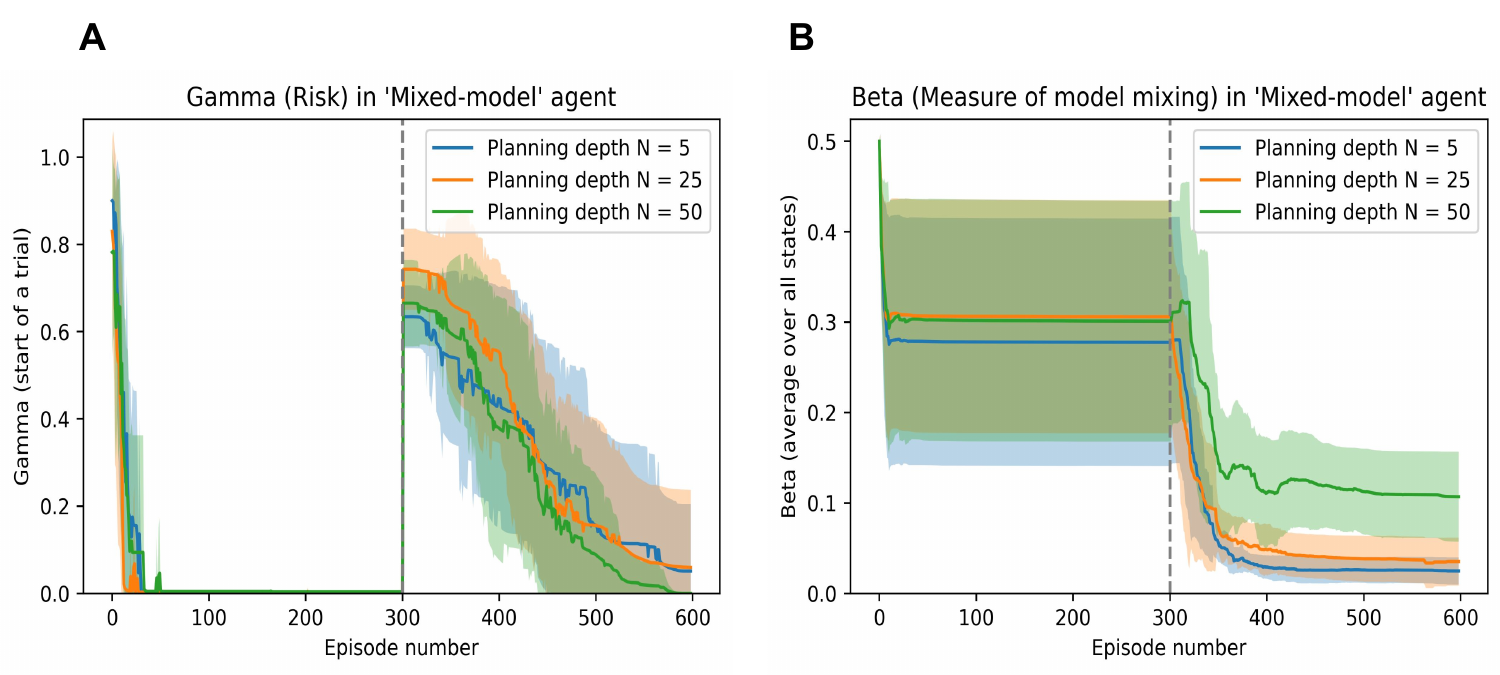}}
  \caption{\textbf{A:} Evolution of the Risk parameter ($\Gamma$) of the mixed-model agent when embodied in the mutating grid scheme, \textbf{B:} Evolution of the model mixing parameter ($\beta$) of the mixed-model agent when embodied in the mutating grid scheme.}
  \label{beta-gamma-evolution-mut}
\end{figure}

An additional advantage of the mixed model proposed (and the POMDP-based generative models) is that we can probe the model parameters to understand the basis of intelligent behaviour demonstrated by agents through the lens of active inference. Models that rely on artificial neural networks (ANNs) to scale up the models \cite{Fountas2020} have limited explainability regarding how agents make decisions, especially when faced with uncertainty.

In Fig.\ref{beta-gamma-evolution-mut}: (A), we can probe to see the evolution of the risk ($\Gamma_t$) in the model (associated with the CL method scheme as defined in \cite{Isomura2022a}).
We can observe that the model's risk quickly tends to zero when the easy grid is presented and solved; however, it shoots up when faced with the environment mutation.

Similarly, the evolution of the bias parameter (that balances the DPEFE and CL method in the mixed model) is shown in Fig.\ref{beta-gamma-evolution-mut}: (B). Here, we also observe how the agent consistently maintains a higher bias to the DPEFE model when it has a higher planning ability (i.e. the agent with a planning depth of $N=50$ compared to bias in agents with $N = 25$, and $N=5$).
 
We should note that the value of the bias parameter never increases more than 0.5, even when the DPEFE agent is planning at $T=50$. In the simulations, we start with a bias $\beta = 0.5$ and update $\beta$ according to \eqref{beta-update-eqn}. This shows how the agent eventually learns to rely on the mixed model's CL scheme (i.e., experience). Still, the DPEFE component (i.e. planning) accelerates learning and performance to aid decision-making. Such insights into the explainability of the agent's behaviour via model parameters help study the basis of natural/synthetic intelligence.

\subsection{Conclusions}

This paper compared and contrasted two decision-making schemes in the active inference framework. Observing the pros and cons of both approaches, we examined them on tasks that demand spontaneous (Cart Pole - v1) and strategic (maze) decision-making, thereby testing a hybrid approach. The insights observed in this work will help improve algorithms used for control, given the excitement around using active inference schemes \cite{DaCosta_2022_robotics}.

We leave the detailed analysis of behavioural dependence on parameters and model expansion in more demanding environments to future work. Systematic comparison with ANNs (Artificial Neural Networks) aided models like in the results of \cite{Fountas2020} is also a promising direction to pursue.

\section{Software note}

The grid environment and agents (DPEFE, CL and Mixed-model schemes) were custom-written in Python. All scripts are available at the following link: \url{https://github.com/aswinpaul/aimmppcl_2023}.

\section{Acknowledgments}

AP acknowledges research sponsorship from IITB-Monash Research Academy, Mumbai and the Department of Biotechnology, Government of India. TI is funded by the Japan Society for the Promotion of Science (JSPS) KAKENHI (Refs: JP23H04973 \& JP23H03465) and the Japan Science and Technology Agency (JST) CREST (Ref: JPMJCR22P1). AR is funded by the Australian Research Council (Ref: DP200100757) and the Australian National Health and Medical Research Council Investigator Grant (Ref: 1194910).  AR is affiliated with The Wellcome Centre for Human Neuroimaging, supported by core funding from Wellcome [203147/Z/16/Z]. AR is also a CIFAR Azrieli Global Scholar in the Brain, Mind \& Consciousness Program.

\bibliographystyle{unsrtnat}
\bibliography{bib_mm}

\end{document}